\documentclass[conference]{IEEEtran}
\IEEEoverridecommandlockouts
\usepackage{cite}
\usepackage{amsmath,amssymb,amsfonts}
\usepackage{algorithmic}
\usepackage{graphicx}
\usepackage{textcomp}
\usepackage{xcolor}
\def\BibTeX{{\rm B\kern-.05em{\sc i\kern-.025em b}\kern-.08em
    T\kern-.1667em\lower.7ex\hbox{E}\kern-.125emX}}

\usepackage{algorithm}
\usepackage{algorithmic}
\usepackage[switch]{lineno}
 
\usepackage{color}
\usepackage[normalem]{ulem}
\usepackage{rotating}
\usepackage{tabularray}

\newlength\myindent
\newcommand\bindent{%
  \begingroup
  \setlength{\itemindent}{\myindent}
  \addtolength{\algorithmicindent}{\myindent}
}
\newcommand\eindent{\endgroup}

\usepackage[textsize=tiny]{todonotes}

\begin{document}

\title{Kernel-U-Net:  Multivariate Time Series Forecasting \\
  using Custom Kernels 
}

\author{

\hspace{5.0ex}
\and 

\IEEEauthorblockN{ Jiang YOU  \textsuperscript{1, 2, 3} }
\IEEEauthorblockA{
\textit{Université Paris-Est Créteil, }\\
Île-de-France, France\\
jiang.you@esiee.fr}
\and
\hspace{10.0ex}
\and 
\IEEEauthorblockN{Arben CELA \textsuperscript{1, 2, 4} }
\IEEEauthorblockA{
\textit{ESIEE Paris-UGE}\\
Île-de-France, France\\
arben.cela@esiee.fr}
\and
\hspace{10.0ex}
\and 
\IEEEauthorblockN{René NATOWICZ \textsuperscript{1,2} }
\IEEEauthorblockA{
\textit{ESIEE Paris-UGE}\\
Île-de-France, France\\
rene.natowicz@esiee.fr}
\and 
\hspace{13.0ex}
\and
\hspace{2.0ex}
\and
\IEEEauthorblockN{Jacob OUANOUNOU \textsuperscript{3} }

\IEEEauthorblockA{
\textit{HN-Services}\\
Île-de-France, France\\
jouanounou@hn-services.com}
\and
\IEEEauthorblockN{Patrick SIARRY \textsuperscript{1} }
\IEEEauthorblockA{
\textit{Université Paris-Est Créteil}\\
Île-de-France, France\\
siarry@u-pec.fr}
}
\IEEEoverridecommandlockouts \IEEEpubid{\makebox[\columnwidth]{979-8-3503-6813- 0/24/\$31.00~\copyright2024 IEEE\hfill} \hspace{\columnsep}\makebox[\columnwidth]{ }}

\maketitle

{\let\thefootnote\relax\footnote{{
\\\textsuperscript{1} Laboratoire Images, Signaux et Systèmes Intelligents (LISSI), Université Paris-Est Créteil , Île-de-France, France\\\textsuperscript{2} Département Informatique et Télécommunication, ESIEE Paris-Université Gustave Eiffel, Île-de-France, France
 \\\textsuperscript{3} HN-Services, Île-de-France, France
 \\\textsuperscript{4} Artificial Intelligence Laboratory, UMT, Tirana-Albanie \\

\textit{Proceedings of the 18\textsuperscript{th} International Conference On Innovations In Intelligent Systems And Applications}, Craiova, Romania, 2024.
}}}




\begin{abstract}
Time series forecasting task predicts future trends
based on historical information. Transformer-based U-Net architectures, despite their success in medical image segmentation, have limitations in both expressiveness and computation efficiency in time series forecasting as evidenced in YFormer. To tackle these challenges, we introduce Kernel-U-Net, a flexible and kernel-customizable U-shape neural network architecture. The kernel-U-Net encoder compresses the input series into latent vectors, and its symmetric decoder subsequently expands these vectors into output series. Specifically, Kernel-U-Net separates the procedure of partitioning input time series into patches from kernel manipulation, thereby providing the convenience of executing customized kernels. Our method offers two primary advantages: 1) Flexibility in kernel customization to adapt to specific datasets; and 2) Enhanced computational efficiency, with the complexity of the Transformer layer reduced to linear. Experiments on seven real-world datasets, demonstrate that Kernel-U-Net’s performance either exceeds or meets that of the existing state-of-the-art model in the majority of cases in channel-independent settings. The source code for Kernel-U-Net will be made publicly available for further research and
application.
\end{abstract}

\section{Introduction}
Time series forecasting predicts future trends based on recent historical information. It allows experts to track the incoming situation and react timely in critical cases. Its applications range from different domains such as predicting the road occupancy rates from different sensors in the city \cite{Lai_2018}, monitoring influenza-like illness weekly patient cases \cite{wu_autoformer_2021}, monitoring electricity transformer temperature in the electric power long-term deployment \cite{haoyietal-informer-2021} or forecasting temperature, pressure and humidity in weather station \cite{liu2022non} etc.

\begin{figure*}
    \centering
    \includegraphics[width=0.9\textwidth]{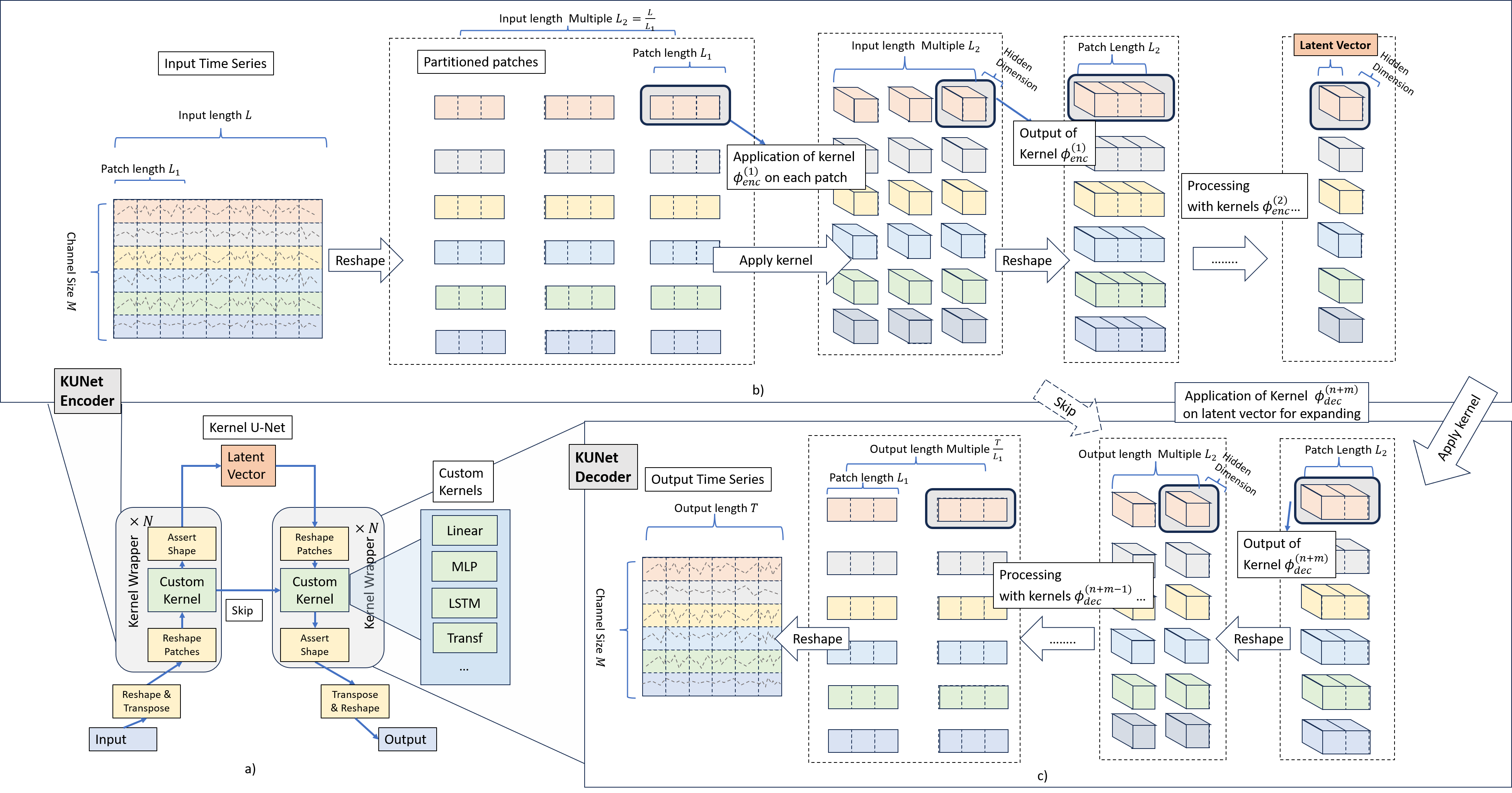} 

    \caption{Illustration of Kernel U-Net Architecture. a) Architecture of Kernel U-Net, it allows executing linear kernel and nonlinear kernels such as MLP, LSTM, and Transformer. b) Illustration of K-U-Net Encoder, the application of $\phi^{(l)}_{enc}$ on patches is independent of the choice of kernel. c) In the K-U-Net decoder, a custom kernel $\phi^{(l)}_{dec}$ expands the vectors into patches in the reverse order. }

    \label{fig:structure_kun_3}
\end{figure*}

Over the past few decades, time series forecasting solutions have evolved from traditional statistical methods\cite{khandelwal_time_2015} and machine learning techniques\cite{persson_multi-site_2017} to deep learning-based solutions, such as recurrent neural networks(RNN) \cite{tokgoz_rnn_2018}, Long Short-term Memory (LSTM) \cite{kong_short-term_2017}, Temporal Convolutional Network (TCN)  \cite{hewage_temporal_2020} and Transformer-based model \cite{li_2020_Enhancing}.

Among the Transformer models applying to time series data, Informer \cite{haoyietal-informer-2021}, Autoformer \cite{wu_autoformer_2021}, and FEDformer \cite{zhou_fedformer_2022} are the best variants that incrementally improved the quality of prediction. As a recent paper \cite{Zeng_AreTE_2022} challenges the efficiency of Transformer-based models with a simple linear layer model NLinear, the authors in \cite{Nie-2023-PatchTST} argued that the degrades of performance comes from the wrong application of transformer modules on a point-wise sequence and the ignorance of patches. By adding a linear patch layer, their model PatchTST successfully relieved the overfitting problem of transformer modules and reached state-of-the-art results. 

We observe that models display distinct strengths depending on the dataset type. For instance, NLinear stands out for its efficiency in handling univariate time series tasks, particularly with small-size datasets. On the other hand, PatchTST is noteworthy for its expressiveness in multivariate time series tasks on large-size datasets. These contrasting attributes highlight the necessity for a unified but flexible architecture. This architecture would aim to integrate various modules easily, allowing for specific customized solutions. Such integration should not only ensure a balance between computational efficiency and expressiveness but also respond to requests for rapid development and testing.

The Convolutional U-net, as a classic and highly expressive model in medical image segmentation\cite{ronneberger_u-net_2015}, features a symmetric encoder and decoder structure that is elegant in its design. This model's structure is particularly suited to the time series forecasting task, as both inputs and outputs in this context are typically derived from the same distribution. The first U-shape model adapted for time series forecasting was the Yformer\cite{Madhusudhanan_yformer_2023}, which incorporated Transformers in both its encoder and decoder components. As mentioned previously, employing Transformers on point-wise data has the potential to cause overfitting issues. Therefore, our investigation aims to discover if there is a U-shape architecture effective in time series forecasting, which also possesses the capability to integrate various modules flexibly, thus facilitating the specific customization of solutions.

To tackle this problem, we propose a flexible and kernel-customizable architecture, Kernel-U-Net (K-U-Net), inspired by convolutional U-net, Swin Transformer, and Yformer for time series forecasting. K-U-Net generalizes the concept of convolutional kernel and provides convenience for composing particular models with non-linear kernels. Following the design pattern, K-U-Net can easily integrate custom kernels by replacing linear kernels with Transformer or LSTM kernels. As a result, K-U-Net can gain expressivity by capturing more complex patterns and dependencies in the data.

Furthermore, the hierarchical structure of K-U-Net exponentially reduces the input length at each level, thereby concurrently decreasing the complexity involved in learning such sequences. Notably, when Transformer modules are utilized in the second or higher-level layers, the computation cost remains linear, ensuring efficiency in processing.

To fully study the performance and efficiency of K-U-Net, we conduct experiments for time series forecasting tasks on several widely used benchmark datasets. We compose 30 variants of K-U-Net by placing different kernels at different layers and then we choose the best candidates for fine-tuning. Our results show that in time series forecasting, K-U-Net exceeds or meets the state-of-the-art results, such as NLinear\cite{Zeng_AreTE_2022} and PatchTST, in the majority of cases. 

In summary, the contributions of this work include:
\begin{itemize}
\setlength\itemsep{0.1em}
\item We propose Kernel-U-Net, a U-shape architecture that progressively compresses the input sequence into a latent vector and expands it to generate the output sequence.
\item Kernel-U-Net generalizes the concept of the convolutional kernel and provides convenience for composing particular models with custom kernels. 
\item The computation complexity is guaranteed in linear when employing Transformer kernels at the second or higher layers.
\item Kernel-U-Net exceeds or meets the state-of-the-art results in most cases.
\end{itemize}

We conclude that Kernel-U-Net stands out as a highly promising option for large-scale time series forecasting tasks. Its hierarchical design provides a balance of low computational complexity and high expressiveness. In most scenarios, it either surpasses or is slightly below the state-of-the-art results. Furthermore, its adaptability in fast-paced development and testing environments is ensured by the use of flexible, customizable kernels.

\section{Related works}
\subsubsection{\textbf{Transformer}}

Transformer \cite{vaswani_attention_2017} was initially introduced in the field of Natural Language Processing (NLP) on language translation tasks. It contains a layer of positional encoding, blocks composed of layers of multiple head attentions, and a linear layer with SoftMax activation. As it demonstrated outstanding performance on NLP tasks, many researchers follow this technique route. 

Vision Transformers (ViTs) \cite{dosovitskiy_image_2021} applied a pure transformer directly to sequences of image patches to classify the full image and outperformed CNN based method on ImageNet\cite{deng_imagenet_2009}.  Swin Transformer \cite{liu_swin_2021} proposed a hierarchical Transformer whose representation is computed with shifted windows. As a shifted window brings greater efficiency by limiting self-attention computation to non-overlapping local windows, it also allows cross-window connection. This hierarchical architecture has the flexibility to model at various scales and has linear computational complexity concerning image size. 

In time series forecasting, the researchers were also attracted by transformer-based models. LogTrans \cite{li_2020_Enhancing} proposed convolutional self-attention by employing causal convolutions to produce queries and keys in the self-attention layer. To improve the computation efficiency, the authors propose also a LogSparse Transformer with only $O(L(log L)^2)$ space complexity to break the memory bottleneck. Informer \cite{haoyietal-informer-2021} has an encoder-decoder architecture that validates the Transformer-like model’s potential value to capture individual long-range dependency between long sequences. The authors propose a ProbSparse self-attention mechanism to replace the canonical self-attention efficiently. It achieves the $O(Llog L)$ time complexity and $O(Llog L)$ memory usage on dependency alignments. Pyraformer\cite{liu2022pyraformer} simultaneously captures temporal dependencies of different ranges in a compact multi-resolution fashion. Theoretically, by choosing parameters appropriately, it achieves concurrently the maximum path length of $O(1)$ and the time and space complexity of $O(L)$ in forward pass. 
PatchTST employs ProbSparse and a linear projection patch layer to reduce the computational complexity of the transformer to $O(\frac {L}{S} \log(\frac {L}{S}))$, where $S$ is patch size. However, its flattened head still incurs a computational cost of $O(\frac{L^2}{S})$.

Meanwhile, another family of transformer-based models combines the transformer with the traditional method in time series processing. Autoformer \cite{wu_autoformer_2021} introduces an Auto-Correlation mechanism in place of self-attention, which discovers the sub-series similarity based on the series periodicity and aggregates similar sub-series from underlying periods. Frequency Enhanced Decomposed Transformer (FEDformer)\cite{zhou_fedformer_2022} captures global properties of time series with seasonal-trend decomposition. The authors proposed Fourier-enhanced blocks and Wavelet-enhanced blocks in the Transformer structure to capture important time series patterns through frequency domain mapping.

\subsubsection{\textbf{U-Net}}

U-Net is a neural network architecture designed primarily for medical image segmentation \cite{ronneberger_u-net_2015}. U-net is composed of an encoder and a decoder. At the encoder phase, a long sequence is gradually reduced by a convolutional layer and a max-pooling layer into a latent vector. At the decoder stage, the latent vector is developed by a transposed convolutional layer for generating an output with the same shape as the input. With the help of skip-connection between the encoder and decoder, U-Net can also capture and merge low-level and high-level information easily.

With such a neat structural design, U-Net has achieved great success in a variety of applications such as medical image segmentation \cite{ronneberger_u-net_2015}, biomedical 3D-Image segmentation \cite{cicek_3d_2016}, time series segmentation \cite{perslev_u-time_2019}, image super-resolution \cite{han_multi-level_2022} and image denoising \cite{zhang_practical_2023}. Its techniques evolve from basic 2D-U-Net to 3D U-Net \cite{cicek_3d_2016}, 1D-U-Net \cite{azar_efficient_2022} and Swin-U-Net \cite{cao_swin-unet_2023} that only use transformer blocks at each layer of U-Net. 

In time series processing, U-time \cite{perslev_u-time_2019} is a U-Net composed of convolutional layers for time series segmentation tasks, and YFormer \cite{Madhusudhanan_yformer_2023} is the first U-Net based model for time series forecasting task. In particular, YFormer applied transformer blocks on each layer of U-Net and capitalized on multi-resolution feature maps effectively.

\subsubsection{\textbf{Hierarchical and hybrid model}}

In time series processing, the increasing size of data degrades the performance of deep models and, crucially, increases the cost of learning them. For example, recurrent models such as RNN, LSTM, GRU have a linear complexity but suffer the gradient vanishing problem when input length increases. Transformer-based block captures better long dependencies but requires $O(L^2)$ computations in general. To balance the expressiveness of complex models and computational efficiency, researchers investigated hybrid models that merge different modules into the network and hierarchical architectures. 

For example, authors in \cite{du_hierarchical_2015} investigated a tree structure model made of bidirectional RNN layers and concatenation layers for skeleton-based action recognition, authors in \cite{xiao_semantic_2016} stacked RNN and LSTM with Attention mechanism for semantic relation classification of text, authors in \cite{kowsari_hdltex_2017} applied hierarchical LSTM and GRU for document classification. In time series processing,  the authors in \cite{hong_multivariate_2017} combine Deep Belief Network (DBN) and LSTM for sleep signal pattern classification.

To meet the demand of balancing the quality of prediction and efficiency in learning Transformer-based models, researchers proposed hierarchical structure in Swin-Transformer \cite{liu_swin_2021}, pyramidal structure in Pyraformer \cite{liu2022pyraformer},  U-shape structure in  Yformer \cite{Madhusudhanan_yformer_2023} or patch layer in PatchTST \cite{Nie-2023-PatchTST}. 

Kernel-U-Net is a U-shape architecture that exponentially reduces the input length at each level, thereby concurrently decreasing the complexity involved in learning long sequences. Kernel-U-Net separates the procedure of partitioning input time series into patches from kernel manipulation, thereby providing the convenience of executing customized kernels. By replacing linear kernels with transformer or LSTM kernels, the model gains enhanced expressiveness, allowing it to capture more complex patterns and dependencies in the data. Notably, when Transformer modules are utilized in the second or higher-level layers, the computation cost remains linear, ensuring efficiency in processing.



\section{Method}

\subsection{Problem Formulation}
Let us note by  $x\in \mathbb{R}^{N\times M}$ the matrix which represents the multivariate time series dataset, where the first dimension $N$ represents the sampling time and the second dimension $M$ is the feature size. Let  $L$ be the length of memory or the look-back window, we denote the historical time series $(x_{t+1,1}, ..., x_{t+L, M})$ (or for short, $(x_{t+1}, ..., x_{t+L}) )$. We also denote the future time series $(x_{t+L+1}, ..., x_{t+L+T})$, where $T$ is the length of future horizon and $t\in[0, N-L-T]$ is the time stamp. 

The time series forecasting task takes a multivariate time series as input and predicts a future series. 
Let $x_t$ be the features at the time step $t$, $L$ the length of the look-back window, and $T$ the future horizon. Given a historical data series $(x_{t+1}, ... ,x_{t+L})$, time series forecasting task predicts the value $(\hat{x}_{t+L+1}, ..., \hat{x}_{t+L+T})$ in the future. Then we can define the basic time series forecasting problem: 
\begin{equation}(\hat{x}_{t+L+1}, ..., \hat{x}_{t+L+T})= f(x_{t+1}, ..., x_{t+L})\end{equation}
where $f$ is the function that predicts the future series based on a historical series.

\begin{algorithm}[tb]
    \label{alg:Kernel_U-Net_Encoder}
    \caption{Kernel U-Net Encoder}
    \textbf{Input}: $L_1$, $M_1$, $\{L_2 $,$ ...$,$L_n$,$M_{2}$,$...$,$M_{m}\}$, $\{D_2$,$ ...$,$D_n$, $D_{n+2}$,$...$, $D_{n+m-1}\}$, $\{\phi^{(1)} $,$... $,$\phi^{(n)}$, $\phi^{(n+2)}$,$...$,$\phi^{(n+m)}\}$, $J_h$, $D_h$
\\
    \textbf{Output}: Instance of Kernel U-Net Encoder 

    \begin{algorithmic} 
    \STATE \# Define the init function :
    \STATE \textbf{def} \_\_init\_\_(inputs):
    \bindent
\STATE $J \in \{L_1, L_2, ..., L_n, $ $M_{2}, ..., M_{m}\}$
    \STATE \# Create first layer 
    \STATE layers = [KW($\phi^1, L_1, M_1, 1, D_2)$]
    \STATE \# Create intermediate layers
 \FOR{ $l$ in $\{2,$ $ ...,$ $n,$  $n+2,$ $..,$ $n+m-1\}$}{

    \bindent
   \STATE 			layers.append(KW($\phi^{(l)},$ $J_l,$ $D_l,$ $1,$ $D_{l\_next}$))
    \eindent
}
\ENDFOR
\STATE 	\# Create last layer
\STATE layers.append(KW($\phi^{(n+m)},J_{n+m},D_{n+m},J_h,D_h$))
\STATE encoder = nn.Sequential (*layers)
\eindent
    \end{algorithmic}
\end{algorithm}

\subsection{Kernel U-Net}
Kernel U-Net (K-U-Net) is a neural network featuring hierarchical and symmetric U-shape architecture. It separates the procedure of partitioning input time series into patches from kernel manipulation, thereby providing the convenience of executing customized kernels (Figure~\ref{fig:structure_kun_3}). More precisely, the K-U-Net encoder reshapes the input sequence into a large batch of small patches and repeatedly applies custom kernels on them until the latent vector is obtained. Later, the K-U-Net decoder expands the latent vector into patches gradually at each layer and obtains a large batch of small patches (Figure~\ref{fig:structure_kun_3}). At last, K-U-Net reshapes the patches to get the final output. In the following paragraphs, we describe methods such as the hierarchical partition of the input sequence and the creation of Kernel-U-Net. 

\subsubsection{\textbf{Hierarchical partition of the input and output}}

In the first place, We split the input trajectory matrix into patches. Let us consider a trajectory matrix $X_t\in\mathbb{R}^{L\times M}$ at time step $t$. Given a list of multiples $\{L_2, ..., L_n\}$ for look-back window and $\{M_2, ..., M_m\}$ for feature, the patch length $L_1$ and feature unit $M_1$ such that $L=\prod_1^n L_k$ and $M=\prod_1^m M_k$. We reshape $X_t$ as a set of small patches $P_t=\{X_{t, i, j}|X_{t, i, j}\in \mathbb{R}^{L_1 \times M_1 }\}$, where $i \in \{1,..., \prod_2^n L_k\}$ and $j\in \{1,..., \prod_2^m M_k\}$. The total number of patches is the product of the multiples of length and feature size $\#P_t =   \prod_2^m M_k \cdot \prod_2^n  L_k $. 

The partitioned patches will be processed by kernels in the encoder gradually and their size will be reduced after the kernel operation at each layer. In the decoder, the patches are generated from vectors of length $1$ at each layer. Since the decoder is symmetrical to the encoder, there will also be a trajectory matrix $\hat{X}_{t+L}\in\mathbb{R}^{T\times M}$ composed of a set of generated patches $\hat{P}_t=\{\hat{X}_{t+L, i, j}\}$ as output. For a simpler description, we let the look-back window $L$ and the forecasting horizon $T$ be equal.

\begin{algorithm}[tb]
    \label{alg:Kernel_Wrapper}
    \caption{Kernel Wrapper (KW)}
    \textbf{Input}: $\phi$, $X$, $J_{in}$,  $D_{in}$, $J_{out}$, $D_{out}$
    
    \textbf{Output}:  Instance of Kernel Wrapper
    
    \begin{algorithmic} 
    \STATE \# Define the init function :
     \STATE \textbf{def} \_\_init\_\_(inputs):
     \bindent
    \STATE \# Initiation of  main kernel function $\phi$ 
	    \STATE $\phi = $ $\phi (J_{in},D_{in},J_{out},D_{out})$
     
	    \STATE \# Holds variables for global operation
	 \STATE   skip\_out = None
	    \STATE skip\_in = None

    \eindent
    \STATE \# Define the forward function :
    \STATE \textbf{def} forward(x):
\bindent
   \STATE   reshape $x$ into (-1, $ J_{in}$,$D_{in}$)     
    \STATE $X = X +$ $\text{skip\_in}$ $\textbf{\text{if}}$ $\text{skip\_in} $ $ \textbf{\text{is}}$ $ \text{not}$ $ \text{None}$ $ \textbf{\text{else}}$ $X$
    \STATE $Z = \phi(X)$ \# assert $Z$.shape is (-1, $J_{out}$,$D_{out}$) 
    \STATE skip\_out = $Z$ \# will be assigned to skip\_in in decoder
    \STATE \textbf{return} $Z$
 \eindent
    \end{algorithmic}
\end{algorithm}

\subsubsection{\textbf{Hierarchical processing with kernels}}
The hierarchical processing of Kernel U-Net consists of compressing an input sequence at the encoding stage and generating an output sequence at the decoding stage. By default, kernels reduce the dimension of input at each layer in an encoder and increase the dimension in a decoder. Let us consider $\mathcal{X}\in \mathbb{R}^{B\times L\times M}$, a batch of trajectory matrix $X_t$, where $B$ is the batch size. We now describe the shape of the intermediate patch before and after the kernel operation.

At the encoder stage, we compress input $\mathcal{X}\in \mathbb{R}^{B\times L\times M}$ into latent vector $\mathcal{Z}$. Let us assume that $D_{h}$ is the unique dimension of the hidden vectors at each intermediate layer and the latent vector to simplify the problem. Firstly, we reshape $\mathcal{X}$ to $(B, 1, \prod_2^n L_k ,L_1,\prod_2^m M_k ,M_1)$ then transpose it to $(B,\prod_2^m M_k ,\prod_2^n L_k ,L_1, 1, M_1)$ and reshape it to $(B \cdot \prod_2^m M_k \cdot \prod_2^n L_k ,L_1,M_1)$ . We denote this vector $\mathcal{P}^{(1)}_{in}$ as a large batch of small patches ready for processing with a kernel. Secondly, the kernel at first layer can now process $\mathcal{P}^{(1)}_{in}$ and outputs a hidden vector $\mathcal{P}^{(1)}_{out}$ in shape $(B \cdot \prod_2^mM_k \cdot \prod_2^n L_k, 1, D_{h})$. After this operation, we reshape the output $\mathcal{P}^{(1)}_{out}$ to $\mathcal{P}^{(2)}_{in}$ of shape $(B \cdot \prod_2^m M_k \cdot \prod_3^n L_k , L_2, D_{h})$ as input for the next layer. Iteratively, the encoder processes all the multiples $\{L_2, ..., L_n, M_2, ..., M_m\}$ in order, and gives finally a batch of latent vector $\mathcal{Z}=\mathcal{P}^{(n+m)}_{out}$ in shape $(B, 1, D_{h})$ (Algorithm~\ref{alg:Kernel_U-Net_Encoder}).

\begin{table*}[ht!]
    
\caption{\label{multivariate-table}Multivariate time series forecasting results with Kernel U-Net. The prediction lengths $T$ are in \{96, 192, 336, 720\} for all datasets. We note the best results in \textbf{bold} and the second best results in \uline{underlined}.}
\centering
\tiny
\begin{tblr}{
  cell{1}{1} = {c=2}{},
  cell{1}{3} = {c=2}{},
  cell{1}{5} = {c=2}{},
  cell{1}{7} = {c=2}{},
  cell{1}{9} = {c=2}{},
  cell{1}{11} = {c=2}{},
  cell{1}{13} = {c=2}{},
  cell{1}{15} = {c=2}{},
  cell{1}{17} = {c=2}{},
  cell{2}{1} = {c=2}{},
  cell{3}{1} = {r=4}{},
  cell{7}{1} = {r=4}{},
  cell{11}{1} = {r=4}{},
  cell{15}{1} = {r=4}{},
  cell{19}{1} = {r=4}{},
  cell{23}{1} = {r=4}{},
  cell{27}{1} = {r=4}{},
  cell{38}{1} = {c=2}{},
  cell{38}{3} = {c=2}{},
  cell{38}{5} = {c=2}{},
  cell{38}{7} = {c=2}{},
  cell{38}{9} = {c=2}{},
  cell{38}{11} = {c=2}{},
  cell{38}{13} = {c=2}{},
  cell{38}{15} = {c=2}{},
  cell{39}{1} = {c=2}{},
  cell{41}{1} = {r=4}{},
  cell{45}{1} = {r=4}{},
  cell{49}{1} = {r=4}{},
  cell{53}{1} = {r=4}{}, 
  vline{2-17} = {1}{},  
  vline{2-3,5,7,9,11,13,15,17} = {-}{},
  hline{1-3,7,11,15,19,23,27,31} = {-}{},
  rowsep=0.12ex , colsep=3.25ex
}
Methods     &     & K-U-Net   & & PatchTST  & & Nlinear &  & Dlinear &  & FEDformer &  & Autoformer &  & Informer &  & Yformer &  \\
Metric &     & MSE  & MAE  & MSE  & MAE  & MSE     & MAE   & MSE     & MAE   & MSE  & MAE   & MSE   & MAE   & MSE & MAE   & MSE     & MAE   \\
\begin{sideways}ETTh1\end{sideways}  & 96  & \textbf{0.355} & \textbf{0.388} & \uline{0.37}   & \uline{0.4}    & 0.374   & 0.394 & 0.375   & 0.399 & 0.376     & 0.419 & 0.449 & 0.459 & 0.865    & 0.713 & 0.985   & 0.74  \\
  & 192 & \textbf{0.388} & \textbf{0.412} & \uline{0.413}  & \uline{0.429}  & 0.408   & 0.415 & 0.405   & 0.416 & 0.42 & 0.448 & 0.5   & 0.482 & 1.008    & 0.792 & 1.17    & 0.855 \\
  & 336 & \textbf{0.407} & \textbf{0.427} & \uline{0.422}  & \uline{0.44}   & 0.429   & 0.427 & 0.439   & 0.443 & 0.459     & 0.465 & 0.521 & 0.496 & 1.107    & 0.809 & 1.208   & 0.886 \\
  & 720 & \textbf{0.430} & \textbf{0.454} & 0.447     & 0.468     & 0.44    & 0.453 & 0.472   & 0.49  & 0.506     & 0.507 & 0.514 & 0.512 & 1.181    & 0.865 & 1.34    & 0.899 \\
\begin{sideways}ETTh2\end{sideways}  & 96  & \textbf{0.269} & \textbf{0.335} & \uline{0.274}  & \uline{0.337}  & 0.277   & 0.338 & 0.289   & 0.353 & 0.346     & 0.388 & 0.358 & 0.397 & 3.755    & 1.525 & 1.335   & 0.936 \\
  & 192 & \textbf{0.332} & \textbf{0.377} & \uline{0.339}  & \uline{0.379}  & 0.344   & 0.381 & 0.383   & 0.418 & 0.429     & 0.439 & 0.456 & 0.452 & 5.602    & 1.931 & 1.593   & 1.021 \\
  & 336 & \uline{0.355}  & \uline{0.400}  & \textbf{0.329} & \textbf{0.384} & 0.357   & 0.400 & 0.448   & 0.465 & 0.496     & 0.487 & 0.482 & 0.486 & 4.721    & 1.835 & 1.444   & 0.96  \\
  & 720 & \uline{0.384}  & \uline{0.435}  & \textbf{0.379} & \textbf{0.422} & 0.394   & 0.436 & 0.605   & 0.551 & 0.463     & 0.474 & 0.515 & 0.511 & 3.647    & 1.625 & 3.498   & 1.631 \\
\begin{sideways}ETTm1\end{sideways}  & 96  & \textbf{0.275} & \textbf{0.331} & \uline{0.29}   & \uline{0.342}  & 0.306   & 0.348 & 0.299   & 0.343 & 0.379     & 0.419 & 0.505 & 0.475 & 0.672    & 0.571 & 0.849   & 0.669 \\
  & 192 & \textbf{0.320} & \textbf{0.361} & \uline{0.332}  & \uline{0.369}  & 0.349   & 0.375 & 0.335   & 0.365 & 0.426     & 0.441 & 0.553 & 0.496 & 0.795    & 0.669 & 0.928   & 0.724 \\
  & 336 & \textbf{0.349} & \textbf{0.380} & \uline{0.366}  & \uline{0.392}  & 0.375   & 0.388 & 0.369   & 0.386 & 0.445     & 0.459 & 0.621 & 0.537 & 1.212    & 0.871 & 1.058   & 0.786 \\
  & 720 & \textbf{0.401} & \textbf{0.412} & \uline{0.416}  & \uline{0.42}   & 0.433   & 0.422 & 0.425   & 0.421 & 0.543     & 0.49  & 0.671 & 0.561 & 1.166    & 0.823 & 0.955   & 0.703 \\
\begin{sideways}ETTm2\end{sideways}  & 96  & \textbf{0.157} & \textbf{0.243} & \uline{0.165}  & \uline{0.255}  & 0.167   & 0.255 & 0.167   & 0.26  & 0.203     & 0.287 & 0.255 & 0.339 & 0.365    & 0.453 & 0.487   & 0.529 \\
  & 192 & \textbf{0.213} & \textbf{0.283} & \uline{0.22}   & \uline{0.292}  & 0.221   & 0.293 & 0.224   & 0.303 & 0.269     & 0.328 & 0.281 & 0.34  & 0.533    & 0.563 & 0.789   & 0.705 \\
  & 336 & \textbf{0.266} & \textbf{0.320} & \uline{0.274}  & \uline{0.329}  & 0.274   & 0.327 & 0.281   & 0.342 & 0.325     & 0.366 & 0.339 & 0.372 & 1.363    & 0.887 & 1.256   & 0.904 \\
  & 720 & \textbf{0.343} & \textbf{0.377} & \uline{0.362}  & \uline{0.385}  & 0.368   & 0.384 & 0.397   & 0.421 & 0.421     & 0.415 & 0.433 & 0.432 & 3.379    & 1.338 & 2.698   & 1.297 \\
\begin{sideways}Electricity\end{sideways} & 96  & \textbf{0.128} & \textbf{0.219} & \uline{0.129}  & \uline{0.222}  & 0.141   & 0.237 & 0.14    & 0.237 & 0.193     & 0.308 & 0.201 & 0.317 & 0.274    & 0.368 & -  & -     \\
  & 192 & \textbf{0.145} & \textbf{0.234} & \uline{0.147}  & \uline{0.24}   & 0.154   & 0.248 & 0.153   & 0.249 & 0.201     & 0.315 & 0.222 & 0.334 & 0.296    & 0.386 & -  & -     \\
  & 336 & \textbf{0.160} & \textbf{0.250} & \uline{0.163}  & \uline{0.259}  & 0.171   & 0.265 & 0.169   & 0.267 & 0.214     & 0.329 & 0.231 & 0.338 & 0.3 & 0.394 & -  & -     \\
  & 720 & \textbf{0.196} & \textbf{0.283} & \uline{0.197}  & \uline{0.29}   & 0.21    & 0.297 & 0.203   & 0.301 & 0.246     & 0.355 & 0.254 & 0.361 & 0.373    & 0.439 & -  & -     \\
\begin{sideways}Traffic\end{sideways}     & 96  & \textbf{0.354} & \textbf{0.229} & \uline{0.36}   & \uline{0.249}  & 0.41    & 0.279 & 0.41    & 0.282 & 0.587     & 0.366 & 0.613 & 0.388 & 0.719    & 0.391 & -  & -     \\
  & 192 & \textbf{0.372} & \uline{0.262}  & \uline{0.379}  & \textbf{0.25}  & 0.423   & 0.284 & 0.423   & 0.287 & 0.604     & 0.373 & 0.616 & 0.382 & 0.696    & 0.379 & -  & -     \\
  & 336 & \textbf{0.388} & \uline{0.270}  & \uline{0.392}  & \textbf{0.264} & 0.435   & 0.29  & 0.436   & 0.296 & 0.621     & 0.383 & 0.622 & 0.337 & 0.777    & 0.42  & -  & -     \\
  & 720 & \textbf{0.430} & \textbf{0.269} & \uline{0.432}  & \uline{0.286}  & 0.464   & 0.307 & 0.466   & 0.315 & 0.626     & 0.382 & 0.66  & 0.408 & 0.864    & 0.472 & -  & -     \\
\begin{sideways}Weather\end{sideways}     & 96  & \textbf{0.142} & \textbf{0.183} & \uline{0.149}  & \uline{0.198}  & 0.182   & 0.232 & 0.176   & 0.237 & 0.217     & 0.296 & 0.266 & 0.336 & 0.3 & 0.384 & -  & -     \\
  & 192 & \textbf{0.187} & \textbf{0.226} & \uline{0.194}  & \uline{0.241}  & 0.225   & 0.269 & 0.22    & 0.282 & 0.276     & 0.336 & 0.307 & 0.367 & 0.598    & 0.544 & -  & -     \\
  & 336 & \textbf{0.238} & \textbf{0.269} & \uline{0.245}  & \uline{0.282}  & 0.271   & 0.301 & 0.265   & 0.319 & 0.339     & 0.38  & 0.359 & 0.395 & 0.578    & 0.523 & -  & -     \\
  & 720 & \textbf{0.308} & \textbf{0.323} & \uline{0.314}  & \uline{0.334}  & 0.338   & 0.348 & 0.323   & 0.362 & 0.403     & 0.428 & 0.419 & 0.428 & 1.059    & 0.741 & -  & -     
\end{tblr}

\end{table*}

At the stage of the decoder, the operations are reversed. We start with $\mathcal{Q}^{(n+m)}_{in}=\mathcal{Z}$ of shape $(B, 1, D_{h})$, a set of input patches to the decoder. We send it to decoder kernel and get an output $\mathcal{Q}^{(n+m)}_{out}$ in shape $(B, M_m, D_{h})$, then we reshape it to $\mathcal{Q}^{(n+m-1)}_{in}$ in shape  $(B \cdot M_m, 1, D_{h })$ for the next kernel. At the end of this iteration over multiples $\{M_{m-1},$ $...,$ $M_2,$ $L_n,$ $...,$ $L_2,$ $L_1\}$, we finally have a set of patches $\mathcal{Q}^{(1)}_{out}$ in the shape  $(B \cdot \prod_2^m M_k \cdot  \prod_2^n L_k, L_1, M_1)$. The last operations are reshaping it into $(B, \prod_2^m M_k , \prod_2^n L_k ,L_1, 1, M_1)$ , transposing it into $(B, 1, \prod_2^n L_k ,L_1, \prod_2^m M_k , M_1)$ and reshaping it into $(B, L, M)$ as final output. 

\subsubsection{\textbf{Kernel Wrapper}}

The Kernel Wrapper (KW) requires necessary parameters such as kernel $\phi$, input patches $X$, output patches $Z$, patches set size $\mathcal{B}$, input patch length $J_{in}$ and dimension $D_{in}$, output patch length $J_{out}$ and dimension $D_{out}$. The kernel wrapper initiates an instance of a given kernel and calls it in the forward function. It reshapes the input patches, executes with the kernel inside, and then checks the output shape.
The wrapper processes $X$ and outputs $Z$ in the encoder, or it processes the sum of $X$ and an encoder output via skip connection, then gives an output in the decoder (Algorithm~\ref{alg:Kernel_Wrapper}).

\subsubsection{\textbf{Formulation of Kernel Operation}}

We add \textit{enc} and \textit{dec} in the index of variables to differentiate their utilization in the encoder and decoder. We denote $ \mathcal{P} = X_{enc}$ and $\mathcal{Q} = X_{dec}$ the set of patches, $l\in \{1,..., n, n+2, ..., n+m\}$ the index of layers, and $\mathcal{B}$ the patches set size. By default, an encoder kernel at layer $l$ receives $\mathcal{P}^{(l)}_{in} \in \mathbb{R}^{\mathcal{B}^{(l)} \times J^{(l)}_{enc\_in} \times D^{(l)}_{enc\_in}}$ and gives  $\mathcal{P}^{(l)}_{out} \in \mathbb{R}^{\mathcal{B}^{(l)} \times 1 \times D^{(l)}_{enc\_out}}$. The decoder kernel at layer $l$ receives   $\mathcal{Q}^{(l)}_{in} \in \mathbb{R}^{\mathcal{B}^{(l)} \times 1  \times D^{(l)}_{dec\_in}}$ and outputs $\mathcal{Q}^{(l)}_{out} \in \mathbb{R}^{\mathcal{B}^{(l)} \times J^{(l)}_{dec\_out} \times D^{(l)}_{dec\_out}}$.
 
We recall that $i^{(l)},j^{(l)}$ are the indices for multiples of length and feature at layer $l$. Given a set of input  $\mathcal{P}^{(l)}_{in}$ at layer $l$, the output $\mathcal{P}^{(l)}_{out}$ of kernel $\phi^{(l)}_{enc} $ in encoder is written : 
\begin{equation}\mathcal{P}^{(l)}_{out, i^{(l)}, j^{(l)}} = \phi^{(l)}_{enc} (\mathcal{P}^{(l)}_{in, i^{(l)}, j^{(l)}}) \end{equation}
The decoder kernel $\phi^{(l)}_{dec} $ at layer $l$ takes the sum of input  $\mathcal{Q}^{(l)}_{in}$ and encoder output $\mathcal{P}^{(l)}_{out}$ via the skip connection as input and produces $\mathcal{Q}^{(l)}_{out}$ as output: 
\begin{equation}\mathcal{Q}^{(l)}_{out, i^{(l)}, j^{(l)}} = \phi^{(l)}_{dec} (\mathcal{Q}^{(l)}_{in, i^{(l)}, j^{(l)}} + \mathcal{P}^{(l)}_{out, i^{(l)}, j^{(l)}})
\end{equation}

Remark that in case of $l=n+m$, we have $\mathcal{Q}^{(l)}_{in, i^{(l)}, j^{(l)}} + \mathcal{P}^{(l)}_{out, i^{(l)}, j^{(l)}}=\mathcal{P}^{(l)}_{out, i^{(l)}, j^{(l)}}$ because there are no higher layers and the kernel only process the encoder output. 

\subsubsection{\textbf{Creation of Kernel U-Net}}

We create the encoder, decoder, and the U-Net in order. The algorithm passes parameters describing the multiples, kernels, and hidden dimensions. Let us note the input length $L$, feature dimension $M$, concatenated lists of multiples of look-back window and feature $\{L_2, $ $..., $ $L_n, $  $M_2, $ $..., $ $M_m\}$, list of hidden dimension of same size $\{D_h,$ $...,$ $D_h\}$, patch size $L_1$ and feature unit $M_1$ , a list of kernels $\{\phi^{(1)}_{enc},$ $ \phi^{(2)}_{enc},$ $...,$ $\phi^{(n)}_{enc},$  $\phi^{(n+2)}_{enc},$ $...,$ $\phi^{(n+m)}_{enc}\}$, latent vector length $J_h$ and dimension $D_h$. We also use \textit{next} and \textit{prev} to iterate over the index $l$. We set the hidden dimension of intermediate output vectors within layers to be equal to that of latent vector for simpler description. It corresponds to channel size in a convolutional network and can be augmented for a larger passage of information if necessary. We describe the creation of the K-U-Net encoder in Algorithm~\ref{alg:Kernel_U-Net_Encoder}. 

The decoder is symmetrical to the encoder and applies kernels in reverse order. More precisely, the decoder takes $J_h$ and $D_h$, multiples
$J\in \{L_1,$ $L_2,...,$ $L_n,$  $M_2,$ $...,$ $M_{m}\}$, kernels $\{\phi^{(n+m)}_{dec},$ $...,$ $\phi^{(n+2)}_{dec},$ $\phi^{(n)}_{dec},$ $...,$ $\phi^{(2)}_{dec},$ $\phi^{(1)}_{dec}\}$ and initiates
kernel wrappers  KW($\phi^{(n+m)}_{dec},$ $J_h,$ $D_h,$ $J_{n+m}, D_{n+m})$,  
KW($\phi^{(l)}_{dec},$ $1,$ $D_{l\_prev},$ $J_{l}, D_{l})$,  
KW($\phi^{(1)}_{dec},$ $1,$ $D_2,$ $L_1, M_1)$ at the highest, intermediate, and lowest layers respectively.

The K-U-Net initiates an encoder and a decoder. In the forward function, the encoder processes the input and generates a list of outputs at each layer and a latent vector. It assigns the skip\_out from encoder kernels to   skip\_in in decoder kernels. Then the decoder takes the outputs from the encoder via skip-connection and the latent vector to generate the final result. 

\begin{figure}
    \centering
    \includegraphics[width=0.8\columnwidth]{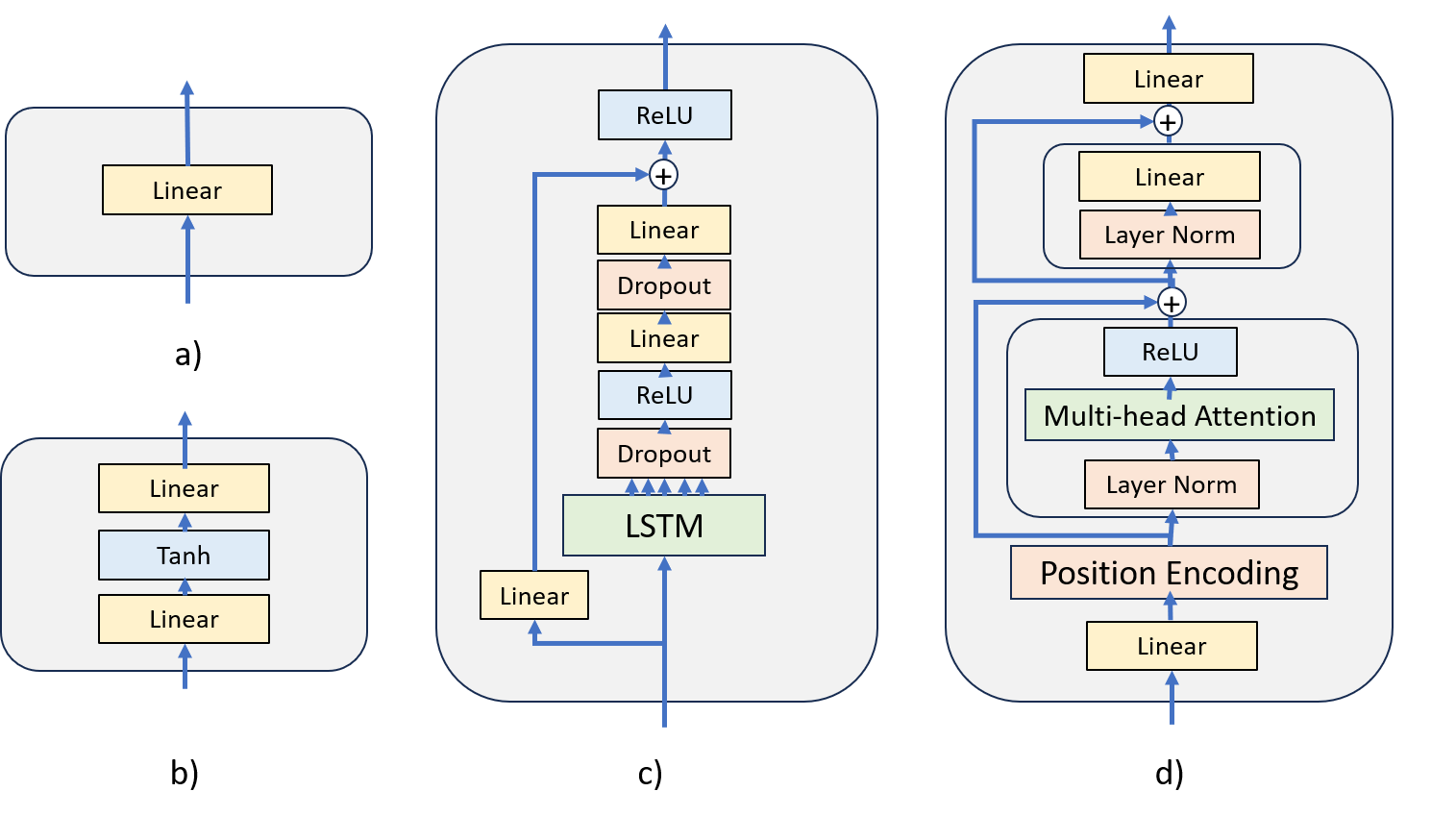} 

    \caption{Structure of Linear Kernel. a) Linear Kernel, b) Multi-layer perceptron (MLP) Kernel with Tanh  activation, c) LSTM Kernel, d) Transformer Kernel}
    
    \label{fig:lienar_kernels}
\end{figure}


\subsection{Custom Kernels}

\subsubsection{\textbf{Linear kernel}}

The linear kernel is a simple matrix multiplication. Given $X\in \mathbb{R} ^{\mathcal{B} \times J_{in} \times D_{in}}$, linear kernel $\phi$ reshape it to $(\mathcal{B}, J_{in} \cdot D_{in})$ and process it as follow: \begin{equation}Z = \phi (X) = X W + b\end{equation}
where $Z$ is output of kernel, $W\in \mathbb{R}^{J_{in}\cdot D_{in}\times J_{out}\cdot D_{out}  }$ is weight matrix and $b\in \mathbb{R}^{J_{out}\cdot D_{out}}$ is bias vector. Remark that the number of parameters of $W$ is $J_{in} \cdot D_{in}  \cdot J_{out} \cdot D_{out}$ and this kernel operation is equivalent to the process with a 1D convolutional layer whose kernel size is $J_{in}$ or $J_{out}$.

\subsubsection{\textbf{Multi-Layer Perceptron kernel}}

The multi-layer perceptron (MLP) kernel has an additional hidden layer and a non-linear activation function Tanh. The formulation is: \begin{equation}Z  = \phi(X) = Tanh(X  W_1 + b_1)  W_2 + b_2\end{equation}
where $Z$ is output of kernel, $W_1\in \mathbb{R}^{J_{in} \cdot D_{in}\times J'\cdot D'  }$ and $W_2\in \mathbb{R}^{J'\cdot D' \times J_{out}\cdot D_{out}}$ are weight matrices , $b_1\in \mathbb{R}^{J'\cdot D'}$ and $b_2\in \mathbb{R}^{J_{out}\cdot D_{out}}$ are bias vectors, and $J' = \frac{1}{2}(J_{in} + J_{out}), D'= \frac{1}{2}(D_{in} + D_{out})$.

\subsubsection{\textbf{Transformer kernel}}

The Vanilla Transformer\cite{vaswani_attention_2017} is made of a layer of positional encoding, blocks that are composed of layers of multiple head attentions, and a linear layer with ReLU activation. The transformer kernel in this work follows the classical structure (Figure~\ref{fig:lienar_kernels}).

\subsubsection{\textbf{LSTM kernel}}

The LSTM kernel contains a classic LSTM cell\cite{long_short_shmiduber} and a linear layer for skip connection. The hidden states of all time steps are combined with a linear layer in the next (Figure~\ref{fig:lienar_kernels}).

\subsection{Complexity Analysis}
Since the K-U-Net is symmetric, we only study the complexity of the encoder layer and assume that the feature size is $1$. Let us suppose that a kernel is receiving a sequence of length $L$ where  $L=\prod_1^n L_i$ and all $L_i$ are equal. As the patch size in the first layer of Kernel U-Net encoder is $L_1$, a kernel module will process $\frac{L}{L_1}$ patches of size $L_1$. Therefore, the complexity is $O(\frac{L}{L_1^n}\cdot g(L_1))$ where $g(L_1)$ is the complexity inside the kernel in the function of patch size $L_1$ and $n$ is the index of the layer. In the case of using the linear kernel at the first layer, the complexity is $O(\frac{L}{L_1} \cdot L_1)$= $O(L)$. In the case of using a classic transformer kernel, the complexity is $O(\frac{L}{L_1} \cdot L_1^2)$= $O(L \cdot L_1)$. Let us set $L_1 = log(L)$, the complexity of the application of such a quadratic calculation kernel is $O(Llog(L))$. Moreover, if we apply the transformer kernel starting from the second layer, the complexity is dramatically reduced to  $O(\frac{L}{L_1 \cdot L_2} \cdot L_2^2)$= $O(L)$. Following the same demonstration, the complexity of using  LSTM kernels and MLP starting from the second layer is also bounded by $O(L)$.

\section{Experiments and Results}
\subsubsection{\textbf{Datasets}}

We conducted experiments with our proposed Kernel U-Net on 7 public datasets, including 4 ETT (Electricity Transformer Temperature) datasets\footnote{https://github.com/zhouhaoyi/ETDataset} (ETTh1, ETTh2, ETTm1, ETTm2), Weather\footnote{https://www.bgc-jena.mpg.de/wetter/}, Traffic\footnote{http://pems.dot.ca.gov} and Electricity\footnote{https://archive.ics.uci.edu/ml/datasets/ElectricityLoadDiagrams20112014}. These datasets have been benchmarked and publicly available on\cite{wu_autoformer_2021} and the description of the dataset is available in \cite{Zeng_AreTE_2022}.


Here, we followed the experiment setting in \cite{Zeng_AreTE_2022} and partitioned the data to $[12, 4, 4]$ months for training, validation, and testing respectively for the ETT dataset. The data is split to $[0.7, 0.1, 0.2]$ for training, validation, and testing for the Weather, Traffic, and Electricity datasets.

\subsubsection{\textbf{Baselines and Experimental Settings }}
We follow the experiment setting in NLinear \cite{Zeng_AreTE_2022} and take $L$ step historical data as input then forecast $T$ step value in the future where $L\in \mathcal{L} =\{336,720\}$ and $T\in \mathcal{T}=\{96,192,336,720\}$. We replace the last value normalization by mean value normalization for ETT, Electricity, and Weather datasets, and apply instance normalization\cite{ulyanov2017instance} for the Traffic dataset. We use Mean Squared Error (MSE) and Mean Absolute Error (MAE) for evaluation as mentioned in \cite{wu_autoformer_2021}.

We include recent methods: PatchTST\cite{Nie-2023-PatchTST}, NLinear, DLinear\cite{Zeng_AreTE_2022}, FEDformer \cite{zhou_fedformer_2022}, Autoformer \cite{wu_autoformer_2021}, Informer  \cite{haoyietal-informer-2021}, LogTrans \cite{li_2020_Enhancing} Yformer\cite{Madhusudhanan_yformer_2023}. We merge the result reported in \cite{Nie-2023-PatchTST} and \cite{Zeng_AreTE_2022} for taking their best in a supervised setting and execute Yformer with default parameters in its Github\footnote{https://github.com/18kiran12/Yformer-Time-Series-Forecasting}.

\subsubsection{\textbf{Experiment details}}

We use a 4-layer K-U-Net for experiments. The list of multiples are respectively \{4, 3, 7\} and \{6, 6, 5\} for look-back windows $L\in \{336, 720\}$. The bottom patch length is 4 and its width is 1. We reshape the input into $(B \cdot M, L, 1)$ for processing with K-U-Net as we follow the channel-independent setting \cite{Zeng_AreTE_2022} and then reshape it back. The hidden dimensions are 128 for multivariate tasks. The learning rate is selected in $[0.00005, 0.001]$. The training epoch is 50 and the patience of early stopping is 10 in general. We use MAE as the loss function.

\subsubsection{\textbf{Model Variants Search}}

We propose 4 kernels for experiments with K-U-Net on 7 datasets. By replacing a linear kernel at different layers with other types of kernels, we search for variants that adapt the dataset. As there are too many variants to name, we note them as "KUN\_$<$kernel$>$\_$<$replace\_layer$>$ ($<$look-back\_window$>$)". For example, KUN\_Linear$\_0000$ $(720)$ means that the model is made of linear kernels at all layers and the look-back window size is 720, KUN\_Transf$\_0100$ means that a transformer kernel replaces the linear kernel at the second layer, KUN\_Linear$\_0110$ means that multilayer perceptron kernels replace the kernels at the second and third layers. We have composed 16 variants with MLP kernels and 7 variants with Transformer and LSTM kernels. 
    

\begin{figure}
    \centering
    \includegraphics[width=0.8\columnwidth]{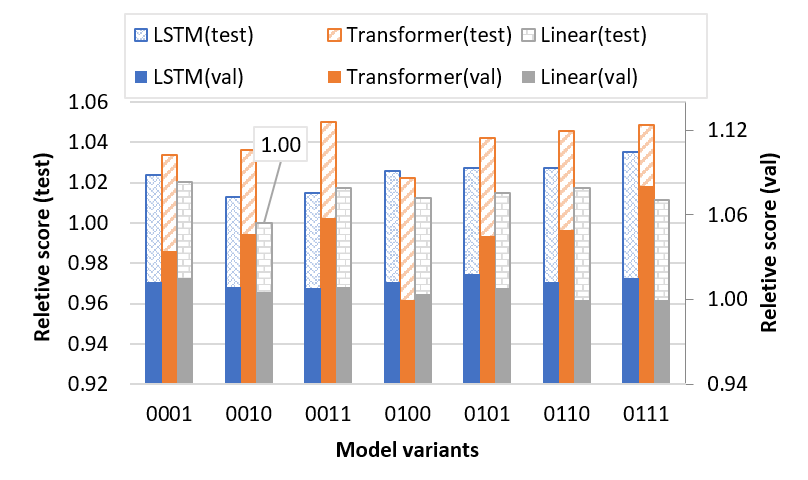} 

    \caption{Search result of K-U-Net with Linear, MLP, LSTM, Transformer kernels on ETTh1 dataset.}
    
    \label{fig:search_etth1_linear_lstm_transformer_v2}
\end{figure}

\begin{figure}
    \centering
    \includegraphics[width=0.8\columnwidth]{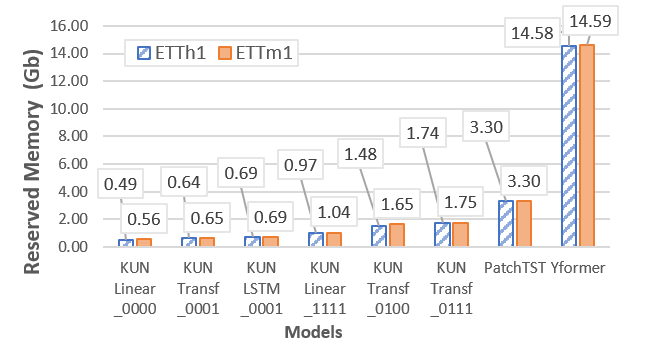} 

    \caption{GPU consumation of K-U-Net, PatchTST and Yformer.}
    
    \label{fig:gpu_consumation_v2}
\end{figure}

To enumerate all variants that achieve at least once the best result, we report their performance by the average of the top 5 minimum running MSE (Top5MMSE) values on the validation and test set. The search results are reported with a relative score(RS) to the minimum Top5-MMSE:
\begin{equation}\text{RS}(V) = \text{Min}_{T\in \mathcal{T}}\frac{\text{Top5MMSE}(V, T)}{\text{Min}_{\{U \in \mathcal{V}, L \in \mathcal{L}\}} \text{Top5MMSE}(U, T, L)}\end{equation}
, where $U$, $V$ are variant models in the models set $\mathcal{V}$, $T$ is a forecasting horizon in $\mathcal{T}$, $L$ is a look back window in $\mathcal{L}$. Relative score notes the best-performed model with 1 and thus helps to identify the high-potential candidates for further fine-tuning examinations.

\begin{figure}
    \centering
    \includegraphics[width=0.8\columnwidth]{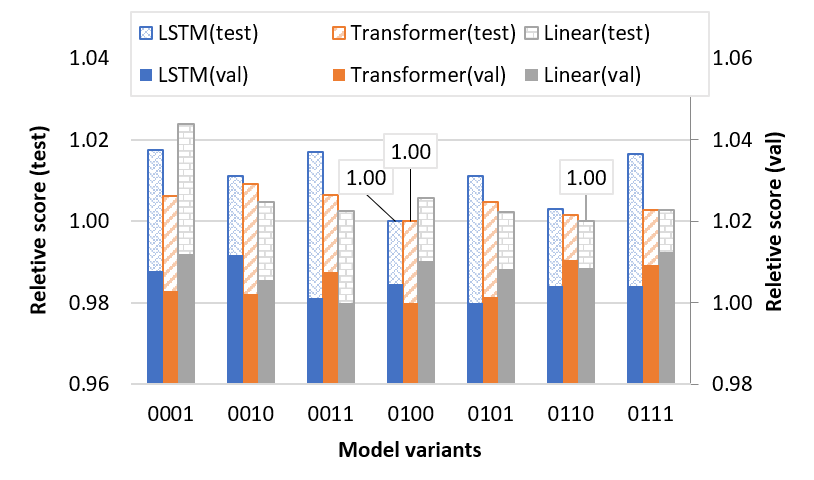} 

    \caption{Search result of K-U-Net with linear, MLP, LSTM, Transformer kernels on Weather dataset.}
    
    \label{fig:search_weather_linear_lstm_transformer_v2}
\end{figure}

We observe in Figure~\ref{fig:search_etth1_linear_lstm_transformer_v2} that the best variant for ETTh1 dataset is KUN\_Linear$\_0010$. Comparing the relative score of the K-U-Net of replace$\_$layer code ($0010$, $0011$),  ($0110$, $0111$) and ($0100$, $0101$), we remark that replacing the highest layer with Transformer and LSTM kernel degrades the performance because of overfitting. Furthermore, We observe in  Figure~\ref{fig:search_weather_linear_lstm_transformer_v2} that the best variants for the Weather dataset are KUN\_Linear$\_0110$, KUN\_LSTM$\_0100$ and KUN\_Transf$\_0100$. Comparing the relative score of the K-U-Net of replace$\_$layer code ($0010$, $0110$),  ($0011$, $0111$) and ($0001$, $0101$) we remark that replacing the second layer with Transformer and LSTM kernel gains the performance for their expressiveness. 

Among candidates in the search phase, we empirically choose 3 variants for fine-tuning experiments with 5 runs. The final result shows that the performance of Kernel U-Net exceeds or meets the state-of-the-art methods in  multivariate settings in most cases.

\subsubsection{\textbf{Multivariate time series forecasting result}}

We remark that our model improves the MSE performance around $72.92\%$ compared with Yformer and  $2.99\%$ compared to PatchTST and NLinear in the multivariate setting (Table \ref{multivariate-table}). It is worth noting that K-U-Net achieves similar results on the Electricity dataset with a variant based on MLP kernels.

\begin{figure}
    \centering
    \includegraphics[width=0.8\columnwidth]{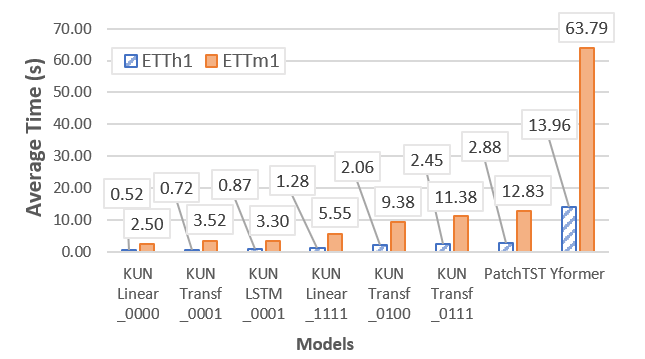} 

    \caption{Computation time of K-U-Net, PatchTST and Yformer.}
    
    \label{fig:computation_time_v2}
\end{figure}

\subsubsection{\textbf{Computation Efficiency}}

 We examined the computation efficiency of 6 variants of K-U-Net, PatchTST, and Yformer. We execute the models on the ETTh1 and ETTm1 datasets for 10 epochs and measure the average execution time per epoch and the GPU consumption during the training. The hidden dimension is set to 128 for all models. For fair comparison, PatchTST and Yformer contain 2 layers of Transformer block which equals to KUN\_Transf$\_0100$. All experiments are executed on a Tesla V100 GPU in a Google Colab environment. Comparing to the PatchTST,  KUN\_Linear$\_0000$ saves $85.20\%$ and $83.07\%$ memory (Figure~\ref{fig:gpu_consumation_v2}) and saves $81.82\%$ and $80.55\%$ computation time (Figure~\ref{fig:computation_time_v2}) on ETTh1 and ETTm1 datasets, KUN\_Transf$\_0100$ saves $55.15\%$ and $50.00\%$ memory and saves $28.47\%$ and $26.88\%$ computation time respectively. 

\section{Conclusion}
In this paper, we propose Kernel-U-Net, a highly potential candidate for large-scale time series forecasting tasks. It provides convenience for composing particular models with custom kernels, thereby it adapts well to particular datasets. As an efficient architecture, it accelerates the procedure of searching for appropriate variants. Kernel-U-Net either exceeds or meets the state-of-the-art results in most cases. In the future, we hope to develop more kernels and hope that Kernel U-Net can be useful for other time series tasks such as classification or anomaly detection.

\bibliography{kun.bib}
\bibliographystyle{nips}


\end{document}